\documentclass[letterpaper, 10 pt, conference]{ieeeconf}  

\IEEEoverridecommandlockouts                              

\usepackage{amsmath} 
\usepackage{amssymb}  
\usepackage{graphicx}
\usepackage{multirow}
\usepackage{rotating}
\usepackage{bbm}
\usepackage{booktabs}
\usepackage{combelow}
\usepackage{epsfig}
\usepackage{graphicx}
\usepackage{pifont}
\usepackage{subfig}
\usepackage{times}
\usepackage{xcolor}
\usepackage[hidelinks]{hyperref}

\title{\LARGE \bf
Dusk Till Dawn: Self-supervised Nighttime Stereo Depth Estimation using Visual Foundation Models 
}

\author{Madhu Vankadari, Samuel Hodgson, Sangyun Shin$^*$, Kaichen Zhou$^*$, Andrew Markham, and Niki Trigoni
\thanks{$*$ refers to equal contribution}%
\thanks{\noindent All authors are with the University of Oxford, Oxford, UK}
}

\begin{document}

\maketitle
\thispagestyle{empty}
\pagestyle{empty}


\begin{abstract}
    Self-supervised depth estimation algorithms rely heavily on frame-warping relationships, exhibiting substantial performance degradation when applied in challenging circumstances, such as low-visibility and nighttime scenarios with varying illumination conditions. Addressing this challenge, we introduce an algorithm designed to achieve accurate self-supervised stereo depth estimation focusing on nighttime conditions. 
    Specifically, we use pretrained visual foundation models to extract generalised features across challenging scenes and present an efficient method for matching and integrating these features from stereo frames.
    Moreover, to prevent pixels violating photometric consistency assumption from negatively affecting the depth predictions, we propose a novel masking approach designed to filter out such pixels.
    Lastly, addressing weaknesses in the evaluation of current depth estimation algorithms, we present novel evaluation metrics. Our experiments, conducted on challenging datasets including Oxford RobotCar and Multi-Spectral Stereo, demonstrate the robust improvements realized by our approach. Code is available at {\color{red}\url{https://github.com/madhubabuv/dtd}}
\end{abstract}

\section{Introduction}

Depth estimation is a pivotal subject within computer vision, with wide-ranging implications for applications such as autonomous driving, augmented and virtual reality, and robotics~\cite{ranftl2020towards, ranftl2021vision}. Despite the accomplishments of supervised depth estimation algorithms, these methods typically depend on high-resolution ground truth data - a challenge that requires substantial computational resources, costly 3D LiDAR sensors, and heavy computational requirements~\cite{godard2017unsupervised, godard2019digging}. 

Addressing the need for ground-truth data, recent research has shown interest in self-supervised depth estimation methods \cite{wang2023sqldepth, peng2021excavating}. Such approaches typically warp the source frame to the target frame using the learned depth information. A photometric loss is found between the reconstructed and actual target images to constrain the learning process. While monocular depth estimation algorithms are widely applicable, they often lack scale information~\cite{zhou2022devnet} and exhibit limited generalizability~\cite{feng2022disentangling}. In contrast, self-supervised stereo depth estimation algorithms that use the correspondence between left and right frames yield more robust performance~\cite{wang2019unos}.

\begin{figure}
  \centering 
  \includegraphics[width=0.45\textwidth]{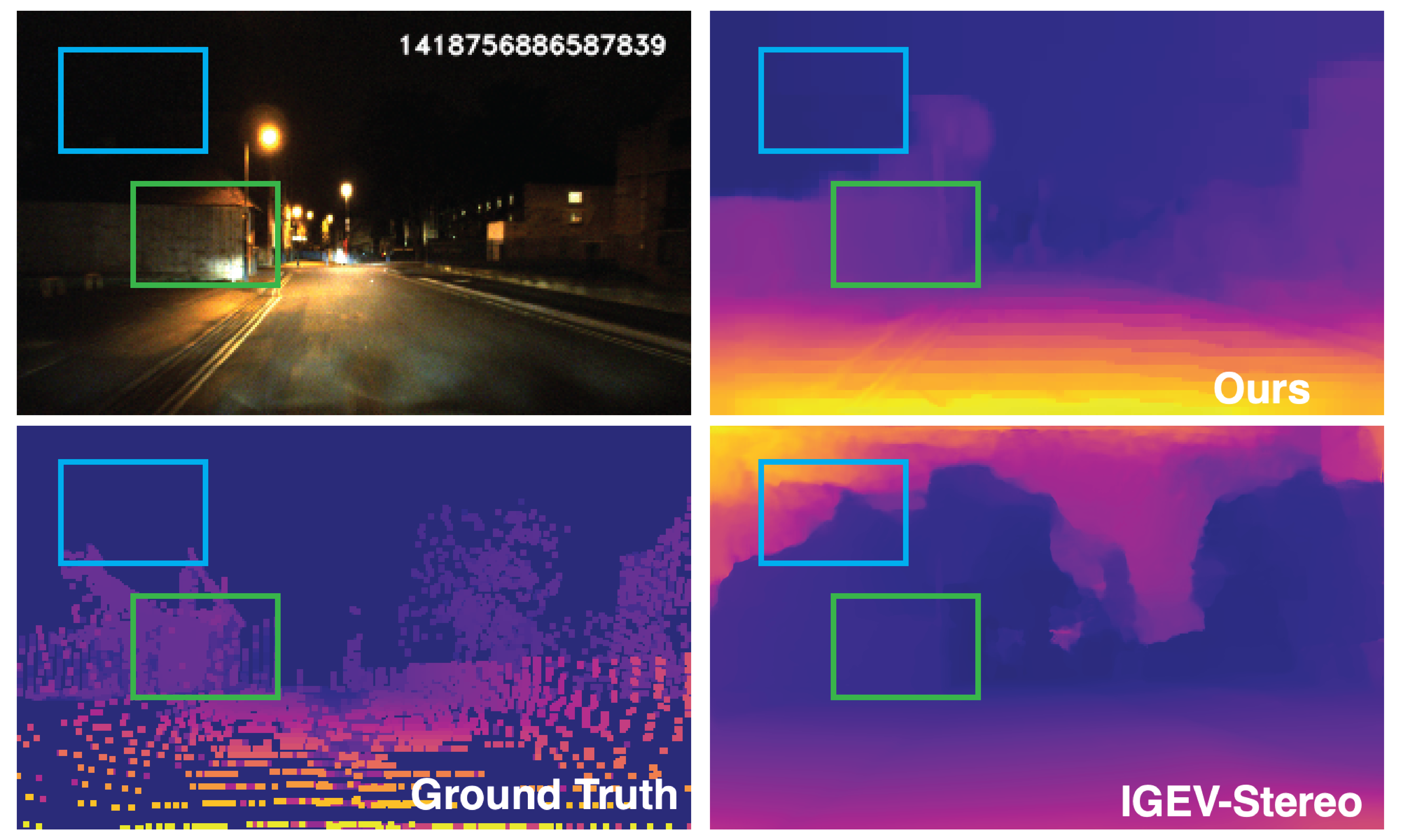}
  \caption{\small A comparison of the estimated disparity using our method (Ours) with a SOTA stereo-matching method, IGEV-Stereo~\cite{xu2023iterative}. Note how the sky (blue rectangle) is incorrectly estimated by IGEV-Stereo as being very near. Similarly, there is a lack of detail showing the edge of the wall and the lamp-post (green rectangle). In comparison, our method is able to accurately estimate these depths.}
  \label{fig:teaserresults}
\end{figure}
Self-supervised stereo depth estimation, however, relies on photometric consistency assumptions and conventional warping relationships, which are constrained by favorable lighting conditions. These assumptions break down in nighttime scenarios, characterized by low texture and fluctuating illumination \cite{gasperini2023robustmde}. Traditional approaches primarily depend on pretrained classification networks, such as \cite{garg2016unsupervisedcnn}, for feature extraction. The utility of these networks is confined to relatively small datasets due to the need for labeled data, and their results are domain-specific. Contemporary self-supervised feature extractors offer enhanced robustness and clarity in feature mapping, as they do not rely on labeled data \cite{caron2021emerging_dino,oquab2023dinov2}. Conventional stereo depth estimation techniques also tend to combine stereo features indiscriminately, without addressing regions that have correspondence issues \cite{wang2019unos, fang2023es3net}. 


In light of this, our approach concentrates on obtaining accurate self-supervised stereo depth estimation at night. The contributions of our paper can be summarized as follows:
\begin{itemize}

    \item We present an architecture capable of efficient self-supervised stereo depth estimation at night, using visual foundation models and a photometric loss function.
    
    \item We introduce a feature-level mask to mitigate the impact of pixels violating illumination assumptions.

    \item We propose a distance regularizer aimed at enhancing the feature descriptions to estimate accurate depth maps.
    
    \item We provide a more rational set of evaluation metrics to assess the performance of depth estimation.
\end{itemize}


\section{Related Work}

Photometric consistency assumes the scene to be static, free of any noise or occlusions, lambertian and temporally illumination invariant. Various changes in the appearance of scenes during nighttime cause systems trained only for daytime datasets to fail. Approaches are required to deal with the scene lighting issues, such as the lack of the sun as a primary source of light.

\subsection{Nighttime Self-supervised Depth Estimation}

Loosely, we classify previous work as being based either on domain adaptation or image enhancement. Adaptation-based approaches seek to overcome the domain shift between day and night using synthesised data to align day/night or clear/inclement weather image encodings~\cite{saunders2023selfsupervisedmde, gasperini2023robustmde}. They aim to obtain the same features across different conditions. Synthesised data is created using standard datasets (i.e. \cite{RobotCarDatasetIJRR}), and a Generative Adversarial Network (GAN). \cite{Vankadari2020UnsupervisedMD} uses a Monodepth2~\cite{godard2019digging} architecture with an adversarially-trained nighttime image encoder. Features are aligned using a GAN to generate day images from night, and a discriminator to enforce similarity. \cite{Zhao2021UnsupervisedMD} emulates this approach, but trains in the output space as well as the feature space. \cite{liu2021selfsupervisedmde} uses the GAN outputs to train two feature extractors, a day-night invariant feature extractor that forms the backbone for depth prediction, as well as night and day style extractors for training. Outside of our broader categorisation, \cite{spencer2020defeatnet} uses a feature space (rather than image space) contrastive loss to improve domain generalisation beyond that of~\cite{godard2019digging}. Enhancement-based approaches aim to improve performance over day models by taking greater account of scene lighting, for example, by isolating illumination information within the image. \cite{wang2021regularizingnighttimeweirdness} uses a learned image enhancement and adapted masking from~\cite{godard2019digging}. \cite{vankadari2022wsgd} estimates the illumination change between the consecutive nighttime images to relax assumptions of photometric loss for better depth estimation.

\subsection{All-day Self-supervised Depth Estimation}

Concurrent state-of-the-art approaches \cite{saunders2023selfsupervisedmde} and~\cite{gasperini2023robustmde} also consider weather conditions in their approaches. Both adapt~\cite{godard2019digging} for their all-day unified networks, using GANs to augment for both weather and lighting conditions. \cite{saunders2023selfsupervisedmde} uses a semi-augmented warping to minimise GAN-induced inconsistency between consecutive frames, and uses raw inputs for pose estimation to minimise error propagation. Similarly,~\cite{gasperini2023robustmde} uses daytime depth estimation to distill depth knowledge to nighttime using image translation networks. \cite{zheng2023steps} uses a complex, partially adversarial architecture, with a learned image enhancement to estimate illumination and uncertainty, which is then masked from the loss. 

\subsection{Supervised Stereo Depth Estimation}

All approaches mentioned so far are monocular, even if stereo images are used during training. In terms of supervised stereo depth estimation, 
\cite{xu2023unifying} uses a transformer and 4D correlation feature matrix, including an iterative refinement inspired by \cite{teed2020raft}, to derive depth, flow, and disparity.  \cite{li2022practical} uses a coarse-to-fine approach with a hierarchical network and adaptive group correlation for getting fine disparities.
\cite{xu2023accurate} builds an attention-based cost volume to suppress redundant information and enhance matching-related information.
\cite{xu2023iterative} unifies stereo and optical flow approaches based on 2D convolution, avoiding the memory cost of 3D convolution.
\cite{sharma2020nighttimestereo} trains a model using day datasets that have ground truth depth and further uses domain adaptation to work on day to night. Our method does not use any daytime data, domain adaptation, or ground truth depth for training. 



\begin{figure*}
  \centering
  \includegraphics[width=0.85\textwidth]{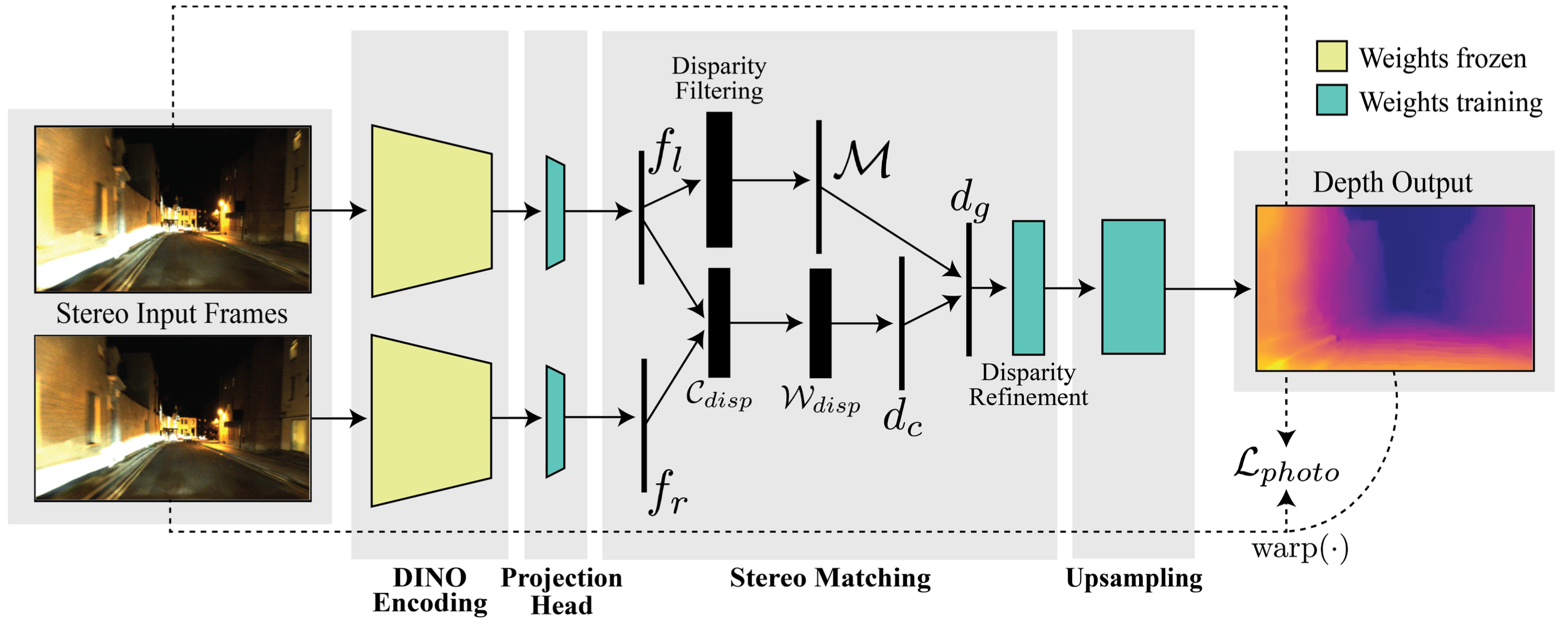}
  \caption{\small Our approach consists of four main elements. Features are encoded independently for each input using DINO \cite{caron2021emerging_dino}, a learnable projection head adapts these features and reduces their dimension, giving $f_{l}$ and $f_{r}$. Stereo matching of the features then takes place, with disparity filtering yielding the mask $\mathcal{M}$, and the combination of $f_{l}$ and $f_{r}$ providing the correspondence volume $\mathcal{C}_{disp}$. $\mathcal{W}_{disp}$ is found by using softmax on $\mathcal{C}_{disp}$, which is used to find coarse disparity $d_{c}$. Coarse disparity and the mask combine to give global disparity $d_{g}$, which is refined and upsampled to give final depth.}
  \label{fig:sysdiagram}
\end{figure*}

\section{Proposed Method}

Given a pair of stereo images $(I_l, I_r)$, we aim to estimate the per-pixel depth map $\mathbf{d}$ using self-supervised learning. The proposed framework is depicted in Fig~\ref{fig:sysdiagram}. Our method is composed of 4 main components, namely the feature extractor, projection head, stereo matcher, and upsampler.

\subsection{Feature Extractor}

In contrast to existing approaches, we use visual foundation models, DINO~\cite{caron2021emerging_dino} and DINOv2~\cite{oquab2023dinov2}, as feature extractors for the input images. The models are trained on the ImageNet~\cite{russakovsky2015imagenet} and Facebook LVD-142M~\cite{oquab2023dinov2} datasets, respectively. In our experiments, we use the pretrained \emph{small}-models of DINO and DINOv2, with patch sizes of $8$ and $14$, respectively. For the rest of the paper, we refer to DINO ViT-S/8 (patch size 8) as the encoder unless otherwise stated. 
Our encoder has a \emph{conv}-layer followed by a series of 12 \emph{transformer} layers. Given an image $I \in \mathbb{R}^{H\times W\times3}$ with height $H$ and width $W$, the conv-layer converts the image into overlapping $8\times 8$ patches with stride $4$, resulting in a feature-map $\hat{f} \in \mathbb{R}^{h'\times w' \times 384}$ where $h' = H/4$, $w' = W/4$. After processing $\hat{f}$  through the first 6 transformer layers, the estimated output feature-map $\Tilde{f} \in \mathbb{R}^{h'\times w' \times 384} $ is subsampled with stride 2 in the height and width dimensions, giving an $8\times$ downscaled feature map that is further processed by 6 more transformer layers resulting $f \in \mathbb{R}^{h\times w\times 384}$, where $h = H/8$, $w = W/8$.

Selection of the transformer layers is based on~\cite{amir2021deep}. They observe that various encoder layers act similarly to hierarchical CNN layers in capturing local-to-global information as the depth of the network increases. Deeper layers capture semantics, and shallow layers capture local details (including positions). Middle layers tend to carry both. For accurate disparity estimation, we use both deep and middle layer features, $(f, \Tilde{f})$, for stereo matching. 

\subsection{Projection Head}

The projection head takes feature vectors $f, \Tilde{f}$ from the encoder and projects them into lower dimensional space $\mathbb{R}^D$ where $D < 384$. This is done because
(1) when PCA is performed on the $384$ dimensions, we observe the 10 first principal components to explain more than $50\%$ of the total variance, suggesting it is safe to project to lower dimensional space and (2) the computational complexity of the stereo matcher during inference and training will be reduced. 

The projection head consists of two \emph{conv}-layers with kernel size as $1\times1$, and $\mathrm{ReLU}$ as activation function in the middle. We use $D = 128$ as the output description dimension and the same projection head for both left and right features.

\subsection{Stereo Matcher}
    \label{sec:stereo_matcher}
    The stereo matcher takes the multi-scale left and right features $\{(f_l,f_r),(\Tilde{f}_l,\Tilde{f}_r)\}$ as input and estimates the disparity map $\tilde{d}_r \in \mathbb{R}^{h'\times w' \times 1}$ in three stages. First, features are enhanced with cross-image feature context using a transformer module. Then, a disparity map $d_g$ is estimated at $1/8$ scale using the $(f_l, f_r)$ features and global matching. Lastly, the disparity is bilinearly interpolated by $2\times$ and corrected for the interpolation artifacts with a residual disparity estimated using $(\Tilde{f}_l,\Tilde{f}_r)$ and local matching. These modules are explained in detail in the following sections.

\noindent\textbf{Transformer:} Stereo images are processed through the encoder and projection head individually, meaning no cross-image interaction or information exchange. In previous works~\cite{xu2023unifying,sun2021loftr,sarlin2020superglue}, it is suggested that adding cross-image information results in better matching accuracy. We therefore use a transformer module similar to~\cite{xu2023unifying} with fixed positional encoding after the projection head. 
To further reduce computational complexity, we only aggregate features that are on the respective epipolar line for any given feature. We use rectified stereo-images that make the search problem 1D, as the epipolar lines are strictly straight. Given $f_l$ and $f_r$, the transformer module outputs $\textbf{f}_l$ and $\textbf{f}_r$ in $R^{h\times w \times D}$.

\noindent\textbf{Coarse Disparity Estimation:} 
The output features from the transformer module are used to compute the dense correspondence volume $C_{disp}$ using normalized feature correlation (i.e, global matching) with a simple matrix multiplication
\begin{equation}
    \mathcal{C}_{disp} = \frac{\textbf{f}_l\cdot\textbf{f}_r^{\top}}{\sqrt{D}} \in \mathbb{R}^{h\times w \times w}.
\end{equation}
We obtain the matching distribution $\mathcal{W}_{disp}$ using a softmax over the last dimension of $\mathcal{C}_{disp}$. This is then multiplied with a 1D pixel grid $\mathcal{P}_{\text{1D}} \in \mathbb{R}^w$ to obtain the corresponding pixel locations $\mathcal{G}_{\text{1D}} \in \mathbb{R}^{h\times w}$. Finally, the coarse disparity $d_c$ can be computed as the difference between $\mathcal{P}_{\text{1D}}$ and $\mathcal{G}_{\text{1D}}$. Formally, this can be written as:
\begin{equation}
    \begin{gathered}
    \mathcal{W}_{disp} = \mathrm{softmax}(C_{disp}),\,\text{and}\,\,\, \mathcal{G}_{\text{1D}} = \mathcal{W}_{disp} \mathcal{P}_{\text{1D}}, \\
    d_c = \mathrm{ReLU}(\mathcal{G}_{\text{1D}} - \mathcal{P}_{\text{1D}}),
\end{gathered}
\end{equation}
where $\mathrm{ReLU}$ ensures the disparity is always positive.
\begin{figure*}[!t]
    \centering
    \includegraphics[width=0.9\textwidth]{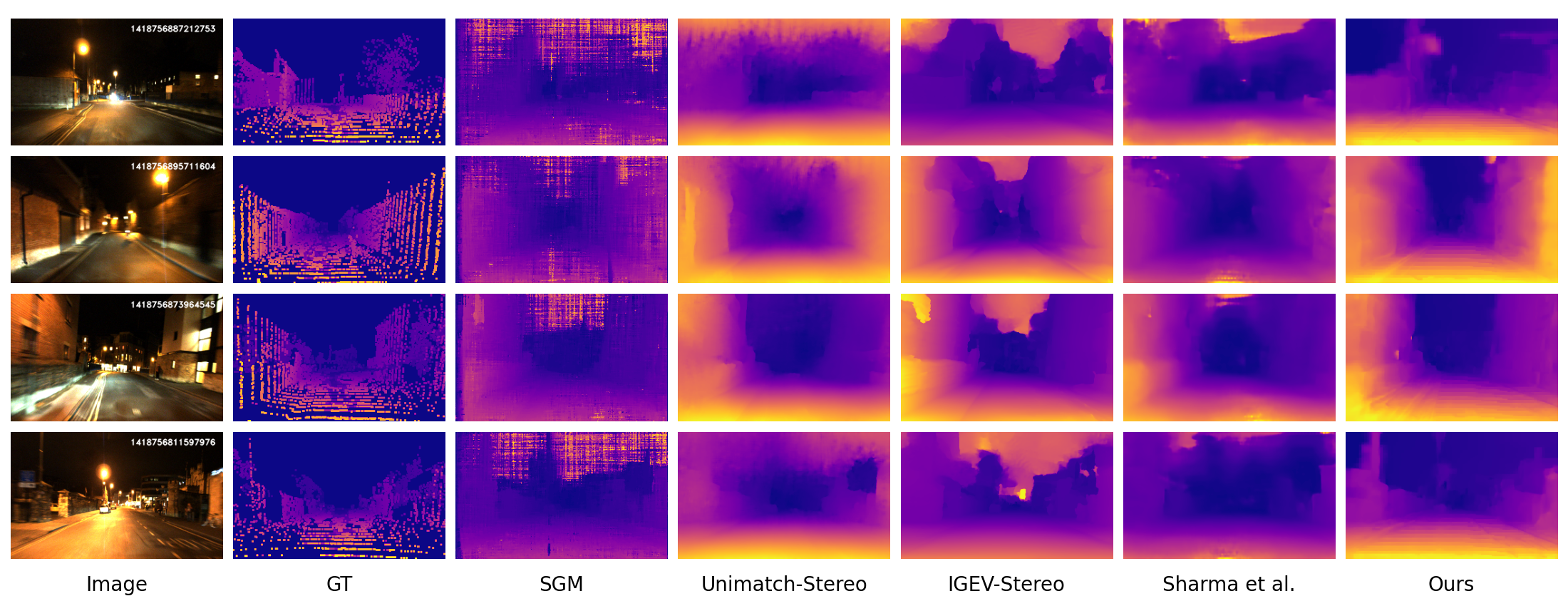}
    \caption{\small The qualitative comparison of the proposed method with SGM~\cite{hernandez2016embedded} and the state-of-the-art supervised methods Unimatch-Stereo~\cite{xu2023unifying}, IGEV-Stereo~\cite{xu2023iterative}, and Sharma et al.~\cite{sharma2020nighttimestereo}. The brighter the pixel is, the closer it is to the camera.}
    \label{fig:depth_qual}
\end{figure*}

    \noindent\textbf{Disparity Filtering:} One observation made during our experiments is that features belonging to areas of low texture, particularly the sky, resulted in very noisy disparity estimates similar to~\cite{hernandez2016embedded, xu2023iterative} as shown in Fig.~\ref{fig:teaserresults}. We explain this as being due to reduced camera sensitivity in low-light causing an accumulation of noise, resulting in erroneous and noisy features. Such features exacerbate incorrect feature matches, causing noisy estimates, as we are extracting disparity as a byproduct, rather than estimating it directly as in~\cite{sharma2020nighttimestereo}. To address this, we propose a simple yet effective solution based on intra-image feature description distances, using them to mask the noisy areas. We conjecture that features belonging to noisy areas tend to have a lower minimum distance from their nearest neighbors, compared to the features that belong to well-lit areas. In a set of $n$ features, given a feature $f_i \in f$ with $i \in \{ 1,2,3...,n \}$: (1) we normalize to have unit length; (2) we estimate one to all cosine similarity to find the nearest neighbor $f_j$, with $j\neq i$; and (3) we calculate $l_2$ distance $p_i$ between $f_i$ and $f_j$. Formally, this is, $f'_j = \mathrm{argmax}_j(f'_i\cdot f'^{\top}_j), \, j\neq i$ where $f' = \frac{f}{\lvert\lvert f \lvert\rvert}$ and  $p_i  = \left |\left| f'_i - f'_j \right| \right |$.
    Estimated distances are then used to filter good disparity values from noisy ones by estimating a mask $\mathcal{M}$, which is used to find masked disparity $d_m$ as:
    \begin{equation}
        d_m = \mathcal{M}\cdot d_c, \quad \mathcal{M} =
    \begin{cases}
            1, &         \text{if } p_i> \zeta,\\
            0, &         \text{otherwise},
    \end{cases}
    \end{equation}
    where $\zeta$ is set to $0.2$ in our experiments.

    \noindent \textbf{Disparity Propagation}
    Some poorly lit regions that contain useful information may get lost during masking. This creates holes in the estimated disparity. We, therefore, propagate the disparity at the valid pixels to the masked-out pixels by measuring self-similarity, as in~\cite{xu2023unifying}. This is done using an attention layer to find global disparity, $d_{g}$:
    \begin{equation}
        d_g = \mathrm{softmax} \left( \frac{f_l\cdot  f_l^{\top}}{\sqrt{D}} \right) \cdot d_m \,
    \end{equation}
    

\noindent\textbf{Disparity Refinement:} The current global disparity $d_g$ is at $1/8 \times$ resolution. To increase the resolution, we upsample $d_g$ using bilinear interpolation by $2\times$. Doing so inevitably introduces interpolation artifacts, so we use fine-level features $\Tilde{f}_l, \Tilde{f}_r$ to account for this. We first warp the right-features $ \Tilde{f}_r$ onto the left frame using the upsampled disparity $d_g^{up}$. The original left-features $\Tilde{f}_l$ and the warped-features $\Tilde{f}'_l$ are then passed through the transformer, disparity estimation, and filtering to estimate residual disparity $\nabla d$. Using localised attention on subsampled features in the refinement process is suggested in \cite{xu2023unifying} to improve accuracy through local feature interactions. The residual disparity is added to the upsampled disparity, giving $d_r = d^{up}_g + \nabla d$. Finally, the refined disparity $d_r$ is correlated with $\Tilde{f}_l$ to output disparity $\tilde{d}_r \in \mathbb{R}^{h'\times w' \times 1}$. 

\subsection{Upsampling}

    The disparity output $\tilde{d}_r$ is upsampled to give final output disparity $d \in \mathbb{R}^{H\times W\times 1}$. Instead of bilinear interpolation, which can result in blurry borders, we use the learnable convex upsampling method proposed by RAFT~\cite{teed2020raft}. In it, two (3 $ \times$ 3) convolutional layers are used to predict a mask and perform softmax over the 9 neighbours of a given pixel.
    

\begin{table*}[!t]
\centering
\scriptsize
\begin{tabular}{cccccccccc}

\toprule
 \multicolumn{1}{l}{\textit{Metric}} &\textit{Method}& type & \textit{Abs.\ Rel.} $(\downarrow)$& \textit{Sq.\ Rel.} $(\downarrow)$& \textit{RMSE} $(\downarrow)$& \textit{Log RMSE} $(\downarrow)$& $\delta < 1.25 (\uparrow)$ & $\delta < 1.25^2 (\uparrow)$ & $\delta< 1.25^3 (\uparrow)$ \\
\midrule


\multirow{3}{*}{U}      & SGM~\cite{hernandez2016embedded} & - & 0.237 &   3.453 & 8.393 &0.358 &   0.689 & 0.862 & 0.924 \\
                        &UniMatch-Stereo~\cite{xu2023unifying} & Sup & 0.207 & 2.521 & 9.087 & 0.373 & 0.588 & 0.793 &0.906 \\
                        &IGEV-Stereo~\cite{xu2023iterative}& Sup & \textbf{0.147} & \textbf{1.655} & 7.092 & 0.312 &  \textbf{0.782} & 0.888 & 0.934 \\
                        &Sharma et al.~\cite{sharma2020nighttimestereo} & Sup & 0.225 &  \underline{1.728} & \textbf{6.489} & 0.278 & 0.669 & \textbf{0.920} & \textbf{0.963} \\
                        &Ours & Self-Sup & \underline{0.177} &  1.970 &  \underline{7.077} &  \textbf{0.274} &  \underline{0.744} & \underline{0.900} &  \underline{0.951} \\

\midrule
\multirow{3}{*}{W}        &SGM~\cite{hernandez2016embedded} & -    & 0.246 & 3.711 & 9.313 & 0.374 & 0.630 & 0.825 & 0.900 \\
                        &Unimatch-Stereo~\cite{xu2023unifying} & Sup & 0.278 & 4.379 & 10.237 & 0.426& 0.422 & 0.660 & 0.828 \\
                        &IGEV-Stereo~\cite{xu2023iterative} & Sup & \textbf{0.184} & 2.649 & 7.433 & 0.327 & \textbf{0.703} & 0.830 &0.894 \\
                        &Sharma et al.~\cite{sharma2020nighttimestereo} & Sup & 0.229 & \textbf{2.113} &  \textbf{6.750} &  \textbf{0.284} & 0.639 &  \textbf{0.892} & \textbf{0.945} \\
                        &Ours& Self-Sup & \underline{0.192} & \underline{2.427} & \underline{7.100} & \textbf{0.275} & \underline{0.703} & \underline{0.870}  & \underline{0.931} \\
\bottomrule
\end{tabular}
\vspace{2mm}
\caption{\small Quantitative evaluation of our proposed method against the SOTA. This evaluation is carried out with a maximum depth range of 50 meters. U refers to the unweighted metric, and W to the proposed weighted metric. In the type-column, ``sup" refers to supervised training, and ``self-sup" refers to self-supervised training. \textbf{Bold} shows the best performance and \underline{underline} refers to second best results.}
\label{tab:depth_quant}
\vspace{-\baselineskip}
\end{table*}

\subsection{Training Losses} \label{sec:losses}

\noindent \textbf{Photometric Loss:} Estimated disparity is used to reconstruct the left image from the right using bilinear interpolation. The reconstructed image is compared with the original left image to calculate the photometric loss. Given the left and right-images $(I_l, I_r)$ and estimated disparity $d$, the photometric loss $L_{\rm photo}$  is:
\begin{gather}
    \mathbf{d} = \frac{b\lambda}{d}, \,\,\, \hat{I}_l = \mathrm{warp}(I_r, \mathbf{d}, K, T_{lr}), \\
    L_{\rm photo} = \alpha \lvert  I_l - \hat{I}_l \rvert  + (1-\alpha) \mathrm{SSIM}(I_l,\hat{I}_l),
\end{gather}
where $\mathbf{d}$ is the output depth map, $b$ is the baseline distance between the left and right cameras, $\lambda$ is the focal length, $K$ is the camera calibration matrix, $T_{lr} \in SE(3)$ is extrinsics of the stereo-rig, $\alpha$ is the convex combination weight between $L_{1}$ and $SSIM$ losses, and is set to $0.15$. The $\mathrm{warp}(\cdot)$ function warps the left from the right image using $\mathbf{d}, K$ and $T_{lr}$. More details of this loss can be found in~\cite{godard2019digging}.

\noindent \textbf{Distance Regularizer:} We also encourage all of the features to maximise the minimum distance from their nearest neighbour using a regularization loss inspired by~\cite{sablayrolles2018spreading}. This allows the features from poorly lit areas to improve. However, there is an imbalance between well and poorly-lit areas in the majority of our training split images, similar to the class imbalance problem from classification literature. This is addressed by using a modulation factor, $\gamma$, to reduce the concentration of loss on features that already have higher minimum distance with their nearest neighbour, and to focus more on small feature distances. Formally:
\begin{equation}
    L_{\rm reg} = -\frac{1}{n} \sum_{i=0}^{n} (1 - p_i)^\gamma \log(p_i),
\end{equation}
where $\gamma$ is a modulation factor similar to focal-loss in~\cite{lin2017focal}, used here to focus more on features that have low minimum distances. We set $\gamma = 2$ in our experiments.

In order to make the estimated disparity spatially smooth while preserving the edges, the common edge-aware Disparity Smoothness Loss $L_{\rm smooth}$ from~\cite{godard2017unsupervised} is used. Finally, the total loss is $L_{\rm total} = L_{\rm photo} + \beta_1 L_{\rm reg} + \beta_2 L_{\rm smooth}$, where $\beta_1$ balances how much we spread the features on the unit-sphere and is set as $\beta_1 = 1$, and we choose $\beta_2 = 0.1$.

\section{Experiments}
\subsection{Datasets}
\label{sec:datasets}
Throughout our experiments, we train on the RobotCar Dataset~\cite{RobotCarDatasetIJRR} and test on both the Robotcar~\cite{RobotCarDatasetIJRR} and MS2~\cite{shin2023deep} datasets. Details of each are given below. 

\noindent\textbf{Oxford RobotCar:} The Oxford RobotCar Dataset~\cite{RobotCarDatasetIJRR} is an autonomous driving dataset collected on the same route over a year in Oxford, UK. 
We follow the data splits proposed in~\cite{vankadari2022wsgd} to exclude the geographical overlaps between training and test splits. We use the six sequences from the traverse on 2014-12-16-18-44-24 for our experiments, providing $19612$ images for training, $4559$ images for validation, and $709$ images for testing.  Ground truth depth data for evaluation is generated by projecting the LiDAR data from several nearby frames into the test frame using the available RTK~\footnote{Quantitative results may change when other forms of pose data such as VO or INS is used to generate ground truth depth} data. 

\noindent\textbf{Multi-Spectral Stereo (MS2) Dataset:}
The MS2 dataset~\cite{shin2023deep} contains 184K pairs taken from multi-spectral sensors on a vehicle in Daejeon, South Korea. 
The sequences include various lighting, weather, and traffic conditions. 
Following the evaluation split proposed in~\cite{shin2023deep}, we use the \textit{Road}, \textit{City}, and \textit{Campus} nighttime sequences, further sub-sampling them with $5\rm m $ distance between consecutive test images. This gave 1,470 pairs for evaluation. We use the (filtered) ground truth depth data released with the dataset for the evaluation.


 
\subsection{Training details}
\label{sec:training}
The framework is trained using the Robotcar dataset for 20 epochs with an input image resolution of $192\times 320$. We used a batch size of $8$ and the Adam~\cite{kingma2015adam} optimizer with a constant learning rate of $1e-4$.

\label{sec:depth_evaluation}
\begin{figure}
    \centering
    \includegraphics[scale=0.16]{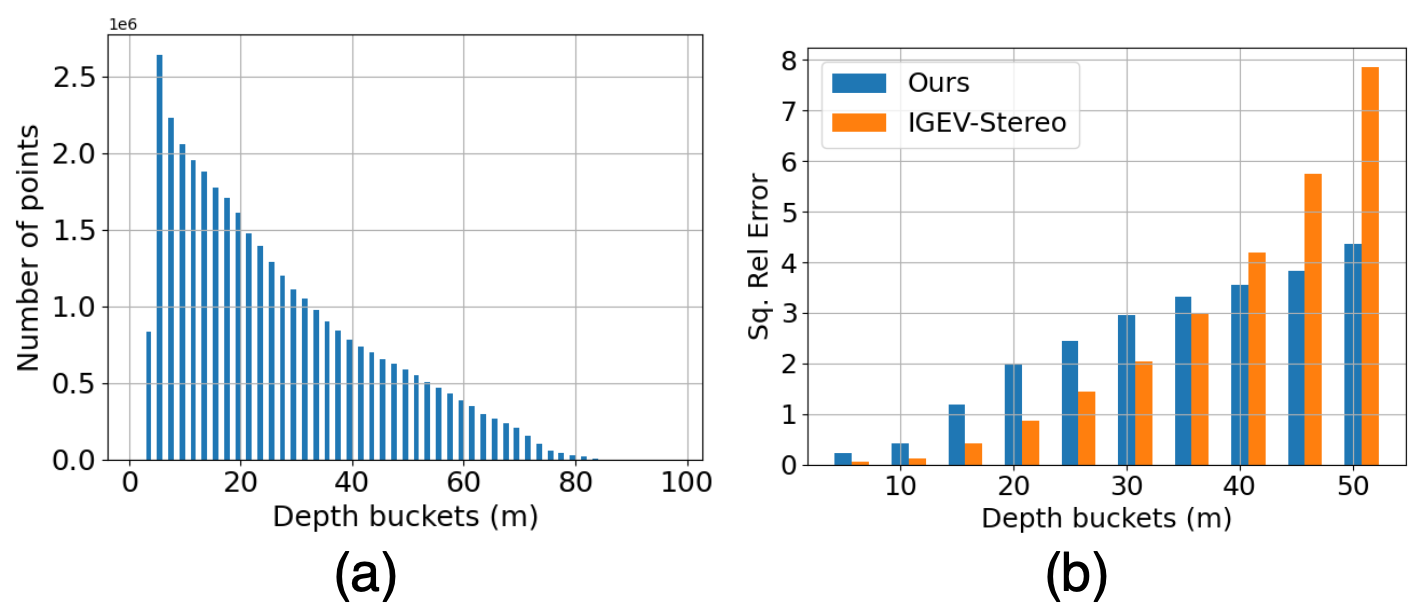}
    \caption{\small The visualization of (a) the ground truth depth distribution of the Robotcar test split, and (b) square relative error calculated at different depth-bins using the proposed weighted metric. }
    \label{fig:new-metric}
\end{figure}

\subsection {Baselines:}
\label{sec:baselines} To the best of our knowledge, there is no self-supervised system that estimates depth from night-time stereo-images. We therefore compare our method with a classical method, Semi-Global Matching (SGM)~\cite{hernandez2016embedded}, and 3 state-of-the-art supervised methods: UniMatch-Stereo\footnote{We used their in-the-wild use stereo-matching with refinement model from GitHub during the evaluation}~\cite{xu2023unifying}, IGEV-Stereo~\cite{xu2023iterative}, and Sharma et al.~\cite{sharma2020nighttimestereo}. Note that these methods are trained end-to-end only for the purpose of stereo-matching, with large amounts of ground truth data. 
Also, we found that the disparity estimation of Unimatch-Stereo and IGEV-Stereo drops drastically when tested at the same resolution as ours. Therefore, we use $2\times$ more resolution while reporting their results. 



\subsection{Depth Evaluation}
\label{sec:depth_eval}
The depth-estimation performance is evaluated qualitatively and quantitatively on both the Robotcar~\cite{RobotCarDatasetIJRR} and MS2~\cite{shin2023deep} datasets.
Quantitative evaluation is carried out using the metrics proposed in~\cite{eigen2014depth}. For every test image, existing methods compute their metrics as the mean of all valid pixels up to a given depth range ($50\rm m$ in our evaluation), usually only with a sparse set of LiDAR points. Taking the overall mean is sensible when pixels are uniformly distributed over the depth range, however, this is not the case for the RobotCar dataset~\cite{RobotCarDatasetIJRR} as shown in Fig~\ref{fig:new-metric}(a). The same effect is observed for most autonomous driving datasets, including the MS2 dataset~\cite{shin2023deep}. This is due to the fact that the majority of pixels in an image are occupied by points that are very close to the camera. Evaluating performance on this kind of data skews the evaluation, as the performance on nearby points outweighs the performance on the far-away points. To mitigate this, we propose the use of depth bins splitting the given depth range into $M$ bins. Letting $\mathrm{x}$ be a metric, unweighted ($U$) and weighted ($W$) evaluation metrics for a given test image can be written as:
\begin{eqnarray}
    \mathrm{x}_{\rm unweighted} = \frac{1}{Z}\sum_{i = 1}^{Z} \mathrm{x}_i, & 
   {\large \mathrm{x}_{\rm weighted} = \frac{1}{M}\sum\limits_{N = 1}^{M}\frac{1}{N_i}\sum\limits_{j = 1}^{N_i} \mathrm{x}_{ij},}
\end{eqnarray}
where $Z$ is the total number of valid pixels, $M$ is the total number of bins and $N_i$ is the total number of valid pixels in the $i$-th depth-bin. We set $M=10$, i.e., each bin covers $5\rm m$ depth. We reported the numbers from both metrics, and we did not use eigen or garg crops~\cite{godard2019digging} during our evaluation.
    \begin{figure}
        \centering 
        \includegraphics[width=0.475\textwidth]{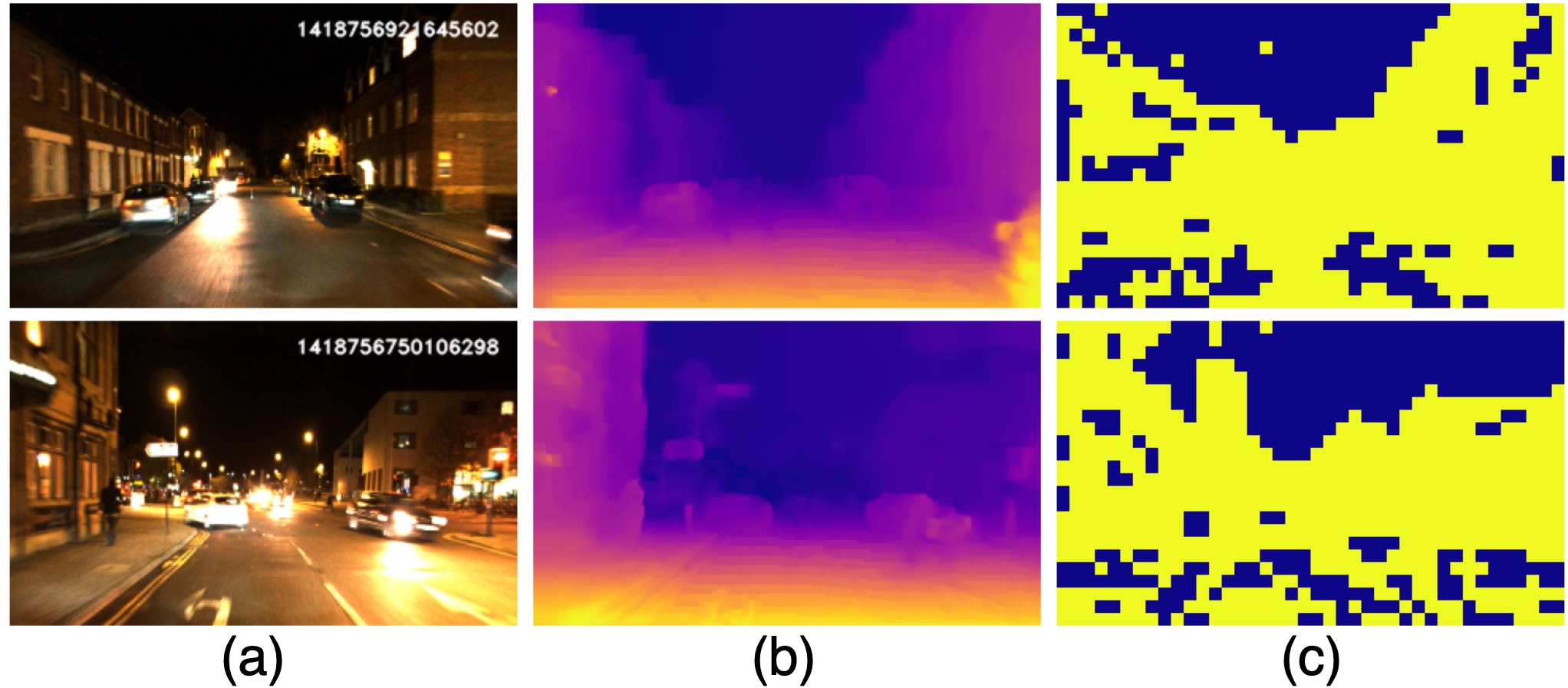}
        \caption{\small Visualization of the estimated masks in (c), with their input-images (Left camera) in (a), and the estimated disparity-maps in (b).}
        \label{fig:noisysky}
    \end{figure}

\noindent\textbf{RobotCar:}  We give a qualitative comparison depicted in Fig~\ref{fig:depth_qual}. Our method is able to extract crisp details, even in poorly illuminated regions. The effect of the masking is clear when looking at the disparity estimated for the sky pixels. We also visualize the estimated masks and the respective filtered disparity maps in Fig.~\ref{fig:noisysky}.
Plausible masks are generated even for noisy low-illumination areas such as the sky. The original ground truth depth images are very sparse, making comparison hard. Hence, we dilated them for visualization. Quantitative results are given in Table~\ref{tab:depth_quant}. Our method performs on par with the baselines in the majority of metrics across both variations despite being self-supervised. Fig~\ref{fig:new-metric}(b) shows the mean squared relative error of IGEV-Stereo~\cite{xu2023iterative} and Ours for the test set. Per the unweighted metrics in Table~\ref{tab:depth_quant}, IGEV-Stereo~\cite{xu2023iterative} performs better. One can see, however, the performance clearly degrades as the depth range increases compared to ours. This effect is much better captured in the weighted metric. Similar observations are made for other metrics as well.  
Lastly, note that our outstanding performance for challenging regions, such as the sky, is not taken into account during the quantitative evaluation due to the absence of the ground truth. As one can see in Fig.~\ref{fig:depth_qual}, all other methods estimate valid depth (brighter pixels) for the sky where they are supposed to be darker, as seen in the ground truth.

\begin{figure}[!t]
    \centering
    \includegraphics[scale = 0.12]{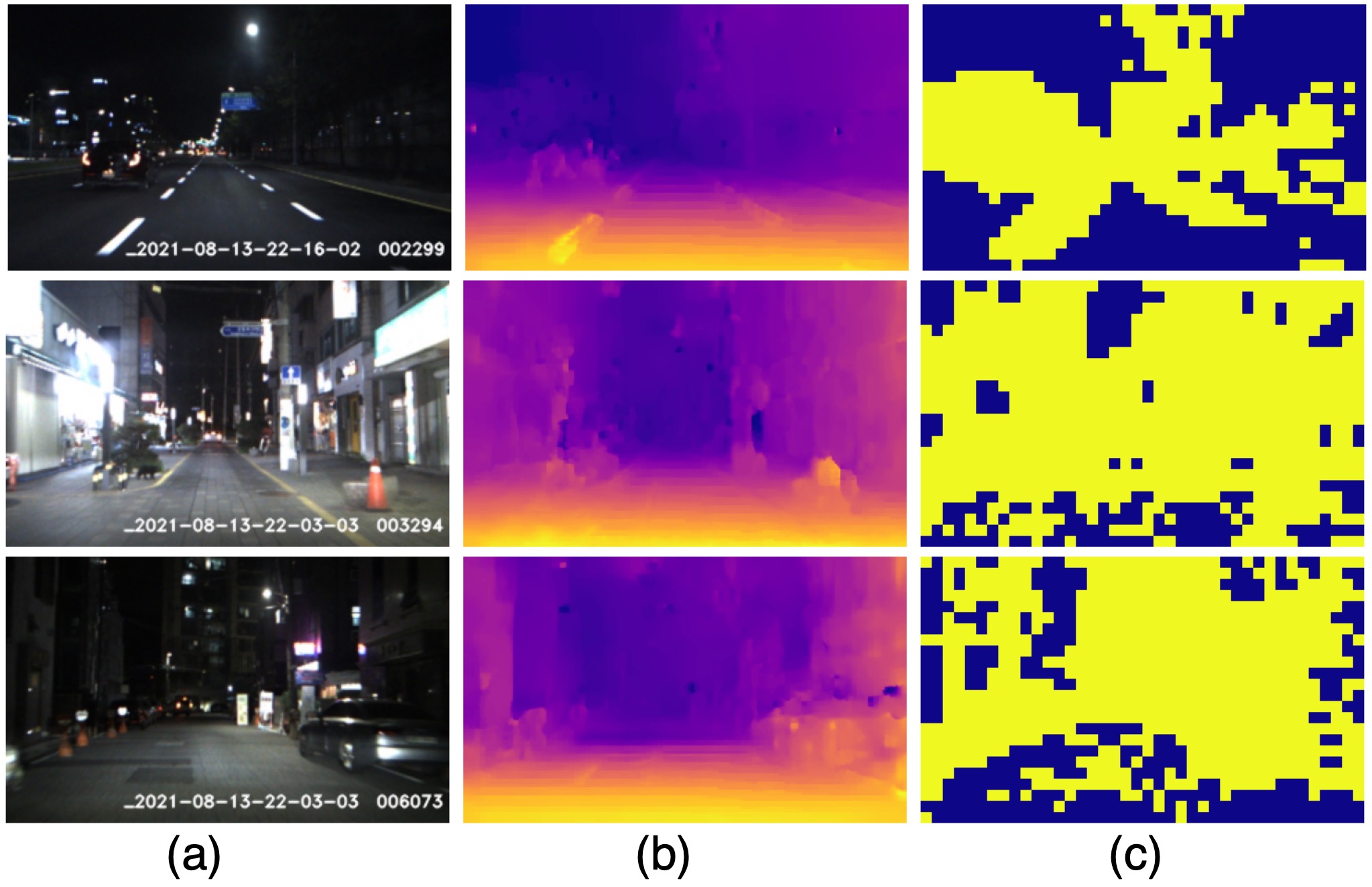}
    \caption{\small Visualization of the estimated disparities in (b) and the masks in (c) with their input-images (Left camera) in (a) when tested on the unseen MS2 dataset.}
    \label{fig:ms2_qual_comp}
\end{figure}

\noindent\textbf{MS2:}
To further evaluate the generalisability of our method, we evaluated the model trained on RobotCar dataset using the test split of the MS2 dataset with the same $50\rm m$ maximum depth range. Despite differences in geographic locations and lighting, and being trained on a relatively small dataset, our model still estimates very plausible disparity maps and pixel masks as shown in Fig~\ref{fig:ms2_qual_comp}. Quantitatively, our method performs better than SGM~\cite{hernandez2016embedded} and is comparable to other methods as shown in Table~\ref{tab:ms2_depth_quant}.  


\begin{table}[!t]
\centering

\scriptsize\addtolength{\tabcolsep}{-1pt}
\begin{tabular}{clllll}
\toprule
\multicolumn{1}{l}{\textit{Metric}} & \textit{Method} & \textit{Abs. \ Rel.} & \textit{RMSE}  & $\delta <$ 1.25 & $\delta<$ $1.25^3$ \\ \hline
\multirow{3}{*}{U}  & SGM~\cite{hernandez2016embedded}   &0.185  & 6.106 &  0.773 & 0.968\\
&Unimatch-Stereo~\cite{xu2023unifying} & 0.095 & 3.396 &  0.910 & 0.994\\
&IGEV-Stereo~\cite{xu2023iterative} & 0.099 & 3.918 & 0.894 &0.985 \\
&Sharma et al.~\cite{sharma2020nighttimestereo} & 0.193 &  5.077 & 0.713 & 0.990 \\
&Ours  & 0.182 & 5.838 &0.740 & 0.977 \\
\midrule

\multirow{3}{*}{W}   &SGM~\cite{hernandez2016embedded}& 0.183  & 6.482 &  0.764 & 0.969\\ 
&Unimatch-Stereo~\cite{xu2023unifying} & 0.112 & 4.182 &  0.860 & 0.991\\ 
&IGEV-Stereo~\cite{xu2023iterative} & 0.125 & 4.977 & 0.815 &0.976 \\
& Sharma et al.~\cite{sharma2020nighttimestereo} & 0.180 & 5.489 & 0.733 & 0.987 \\
&Ours  & 0.180 & 6.162 &0.716 & 0.978 \\
\bottomrule
\end{tabular}
\caption{\small Quantitative evaluation of our proposed method (trained on RobotCar) against the SOTA on MS2 Dataset}
\label{tab:ms2_depth_quant}
\end{table}

\begin{table}
\centering
\scriptsize\addtolength{\tabcolsep}{-3pt}
\begin{tabular}{ccccc}

\toprule
 \textit{Method} & \textit{Abs.\ Rel.} $(\downarrow)$& \textit{RMSE} $(\downarrow)$& $\delta < 1.25 (\uparrow)$ & $\delta< 1.25^3 (\uparrow)$ \\
\midrule

Base model  & 0.201 &  7.734  &  0.724 & 0.943 \\
w/ Mask & 0.214 & 7.644 &  0.712  & 0.948 \\
w/ Mask + Reg  & 0.188 & 7.409 & 0.744  & 0.946 \\
 w/ DINO-V2& 0.204 &   7.871  & 0.711  & 0.944 \\
 \bottomrule
 \end{tabular}
 \caption{\small Ablation study showing the importance of different modules in our system. This evaluation is carried out using the unweighted metrics with 50 meters as the maximum depth.}
\label{tab:ablation}
 \end{table}


\subsection{Ablation Studies}\label{sec: ablation}
We performed various ablation studies on RobotCar dataset to understand the impact of different modules. The results are shown in Table~\ref{tab:ablation}. \textbf{Base model} uses DINO-V1 as the encoder, with stereo-matching and upsampling,  trained using $L_{\rm photo}$ alone. \textbf{W/ mask} had features masked before stereo-matching, also trained with $L_{\rm photo}$ alone. \textbf{W/ mask + reg} used the regularization loss $L_{\rm reg}$ with $L_{\rm photo}$. The improvement in both error and accuracy metrics explains the importance of the proposed masking and regularization loss for accurate depth estimation.  
\subsection{Failure Cases:} Interestingly, training with a DINOv2~\cite{oquab2023dinov2} encoder yielded a performance drop despite being pretrained on a larger amount of data than DINO~\cite{caron2021emerging_dino} as shown in the last row of Table~\ref{tab:ablation}. Also, the overexposed areas and lane markings create undesirable edges in the estimated disparity maps (can be observed in Fig.~\ref{fig:noisysky}, Fig.~\ref{fig:ms2_qual_comp}). This can be a limitation of the commonly used edge-aware disparity smoothness loss $L_{\rm smooth}$. We currently leave these issues for future investigation.

\section{Future Work and Conclusions}
We introduce an algorithm achieving precise self-supervised stereo depth estimation for nighttime conditions, leveraging visual foundation models. We present an efficient masking method and distance regularizer to enhance the accuracy of depth estimation, and novel, weighted evaluation metrics that provide more accurate evaluation given the non-uniform ground truth depth distributions. Our approach shows effective performance across a range of challenging scenarios and generalizes well to unseen datasets. 
\section{Acknowledgments}{This work was supported by AWS via the Oxford-Singapore Human-Machine Collaboration Programme, and EPSRC via ACE-OPS (EP/S030832/1). The authors would also like to thank the anonymous reviewers for their helpful comments.}

\bibliographystyle{IEEEtran.bst}
\bibliography{references}
\end{document}